\documentclass[conference]{IEEEtran}
\IEEEoverridecommandlockouts
\usepackage{cite}
\ifCLASSINFOpdf
\usepackage[pdftex]{graphicx}
\else
\usepackage[dvips]{graphicx}
\fi
\usepackage{amsmath}

\usepackage{amssymb}
\usepackage{amsfonts}
\usepackage{amsopn}
\usepackage{array}
\usepackage{booktabs}
\usepackage[table]{xcolor}
\usepackage[caption=false,font=normalsize,labelfont=sf,textfont=sf]{subfig}
\usepackage[linesnumbered,ruled]{algorithm2e}
\usepackage{url}
\usepackage{microtype}
\usepackage{multirow}
\usepackage{diagbox}

\begin{document}
%
\title{Evolving Deep Convolutional Neural Networks\\ for Hyperspectral Image Denoising}

\author{\IEEEauthorblockN{Yuqiao Liu$^{(a)}$, Yanan Sun$^{(a)}$,
		Bing Xue$^{(b)}$,
		and Mengjie Zhang$^{(b)}$}
	\IEEEauthorblockA{$^{(a)}$College of Computer Science, 
		Sichuan University, Chengdu 610064, China\\ 
		$^{(b)}$School of Engineering and Computer Science, 
		Victoria University of Wellington, Wellington 6140, New Zealand\ 
		\\
		Emails: lyqguitar@gmail.com, ysun@scu.edu.cn, bing.xue@ecs.vuw.ac.nz, and mengjie.zhang@ecs.vuw.ac.nz}
}

\maketitle

\begin{abstract}
Hyperspectral images (HSIs) are susceptible to various noise factors leading to the loss of information, and the noise restricts the subsequent HSIs object detection and classification tasks. In recent years, learning-based methods have demonstrated their superior strengths in denoising the HSIs. Unfortunately, most of the methods are manually designed based on the extensive expertise that is not necessarily available to the users interested. In this paper, we propose a novel algorithm to automatically build an optimal Convolutional Neural Network (CNN) to effectively denoise HSIs. Particularly, the proposed algorithm focuses on the architectures and the initialization of the connection weights of the CNN. The experiments of the proposed algorithm have been well-designed and compared against the state-of-the-art peer competitors, and the experimental results demonstrate the competitive performance of the proposed algorithm in terms of the different evaluation metrics, visual assessments, and the computational complexity.
\end{abstract}

\IEEEpeerreviewmaketitle

\section{Introduction}
Image denoising is one of the fundamental tasks of image processing. Different from the natural 2D image, the Hyperspectral image (HSI) has three dimensions to additionally display the spectral and spatial information. HSIs are widely used in urban planning, agriculture, and forestry~\cite{zhang2011combining,thenkabail2016hyperspectral}. But in the harsh space environment, the multi-detector for generating the HSIs is susceptible, which consequently results in the HSIs having noise. In general, the noise in HSIs has many different types, such as the gaussian noise and the stripe noise. The corrupted hyperspectral data with the noise will affect the accuracy of the consequent work, for instance, the classification tasks~\cite{li2013hyperspectral}. Thus, the HSI denoising has been a hot topic in the past few years\cite{kong2019color,he2018non}. Many algorithms have been proposed, such as the K-singular value decomposition (KSVD)~\cite{elad2006image} and the Tenser-SVD~\cite{zhang2014novel}. Generally, the HSI denoising methods are divided into three different categories as follows.

\noindent{\bfseries 1) Filter-Based Methods:} The core idea of the filter-based methods is to use the filtering operations with a variety of filters including the Fourier transform and the wavelet transform. Particularly, one of the hundreds of channels in an HSI can be regarded as a grayscale image. So, the traditional gray-level image denoising methods, for example, the block-matching 3-D filtering (BM3D)~\cite{dabov2007image}, can be adopted to every channel directly. The limitation of these filtering methods remains in their sensitiveness to the transform function, mainly due to the manually set parameters. In addition, ignoring the correlations across the spectral bands also leads to their relatively poor performance in practice.

\noindent{\bfseries 2) Optimization-Based Methods:} These methods work by adopting reasonable assumptions or the priors, such as the Total Variation (TV), the Non-local (Non-Local), the Sparse Representation (SR), and the Low-Rank (LR) models, etc, focusing on preserving the spatial and spectral characteristics. Because of the high-dimensional feature set and strong spectral correlations in HSIs, the LR regularization has been widely used in HSI denoising, owing to its effective ability of revealing the low-dimensional structure from the high-dimensional data. Due to the high spectral correlation between the adjacent bands and the high similarity in each band, Zhang \textit{et al.}~\cite{zhang2013hyperspectral} proposed the Low-Rank Matrix Recovery (LRMR) for HSIs restoration by transforming a 3-D cube into a 2-D matrix. To better combine the spatial and spectral information, the tensor-based approaches have been presented recently~\cite{fan2018spatial,xie2017kronecker}, which can achieve good results by running on the powerful computation platforms. In summary, although these methods can perform well in HSI denoising, the inadaptability to the mixed noise becomes a barrier to performance improvements.

\noindent{\bfseries 3) Learning-Based Methods:} In recent years, deep learning methods have been proposed and perform well in denoising the HSIs~\cite{chang2018hsi}. The Convolutional Neural Networks (CNNs) are the most representative method of the deep learning-based methods, and widely used in natural image denoising~\cite{jain2009natural,lefkimmiatis2017non,xie2012image}. However, the architecture of CNNs needs to be redesigned as the problem changes. In practice, designing an optimal CNN for the problems at hand is not an easy work, since the best network architecture for a specific problem is unknown, i.e., the depth of the CNN, the number of different types of layers, and the parameters of each layer are are to determine. Meanwhile, the weights of the network which play a vital role in its performance~\cite{sun2019evolving} need to be retrained by a gradient-based algorithm for achieving the promising performance, which highly relies on its initialization~\cite{glorot2010understanding}.

To better explore the merits of CNNs in image denoising, and reduce the human expertise intervention during the architecture design and the weight initialization, in this paper, we propose a novel algorithm (denoted as Evolve-CNN) based on a Genetic Algorithm (GA)~\cite{ashlock2006evolutionary,back1996evolutionary} to design a good architecture and weight initialization for CNNs, to effectively and efficiently address the HSI denoising task. In summary, the contributions of the proposed Evolve-CNN are summarized as follows:

\begin{enumerate}
	\item {\bfseries A novel gene encoding strategy of GA is proposed to encode the individuals for the automated CNN architecture design.} The encoded individual carries the information of the corresponding CNN architecture. A population of such individuals can evolve a better CNN for HSI denoising.
	\item {\bfseries Effective genetic operators are designed for the exploration and the exploitation search.} In GA, the fixed-length chromosome can execute genetic operators easily. However, only a variable-length encoding strategy can search the network architectures efficiently, but there is no accepted good method to play the role of the genetic operators. The proposed operators can perform well on the proposed encoding strategy.
	\item {\bfseries An improved slack binary tournament selection is proposed for choosing promising individuals for offspring generation.} Usually, huge computing resources are required for training CNNs. We regard the complexity as an important evaluation criterion in the environment selection. The improved method can help us search to find a high-performance CNN with lower complexity.
\end{enumerate}

The remainder of this paper is organized as follows. The background of CNN is provided in Section~\ref{related_works}. Section~\ref{proposed_algorithm} describes the proposed algorithm in detail. Section~\ref{experiments} and~\ref{result} provide the experimental design and the experimental results. Finally, the conclusion is given in Section~\ref{conclusion}.

\section{Background}
\label{related_works}
In this section, the skeleton of CNNs is provided, which is the base work of the proposed algorithm. We will introduce the convolution layer, the Reflect Padding (RP) layer, and the Batch Normalization (BN) layer which constitute the skeleton of the CNNs.
\subsection{Convolution Layer}
The convolution layer plays a vital role in CNNs since the convolution operation can extract features from the images. To be specific, the operation provided as the following. First, given an input matrix with the size of n $\times$ n, the filter travels from the top left to the bottom right of the input data to generate a value in the feature map with the convolutional operation. Second, the action is performed again after moving downward with the set stride, until reaching the bottom right of the matrix. Noting that the convolution padding type is an important parameter in CNNs because the type decides the output image size. After a convolution operation where the convolution kernel size is greater than one, the output size will become smaller. There are two padding types in convolution layer: the VALID type and the SAME type. The SAME type can get the same size of the output as that of the input by padding zeros in the input matrix. The VALID type will not pad any elements to the input matrix. Noting that, the VALID type will be used in the proposed algorithm because the SAME type convolutional operation cannot well address the boundary of the images.
\subsection{Reflect Padding (RP)}
\begin{figure}[htp]
	\centering
	\includegraphics[width=0.8\columnwidth]{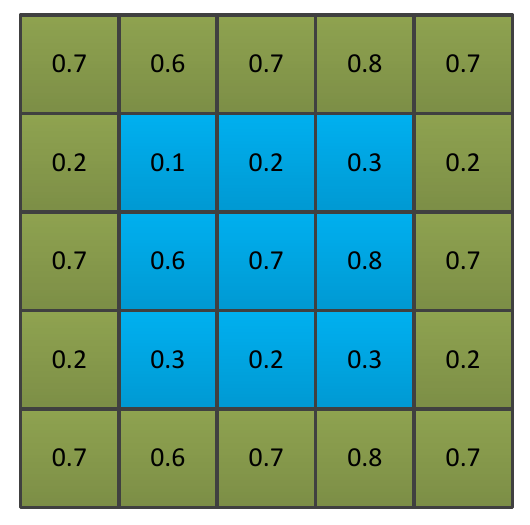}\\
	\caption{An example to show RP. The blue part that represents the input image is given as a $3 \times 3$ matrix, and we use RP to add the green part to get a $5 \times 5$ matrix which represents the original image. The middle of the first row in green is the same as the second row in blue. The other three sides were obtained in the same way. Elements in the green corner are generated by diagonal reflection.}\label{fig_reflect_padding}
\end{figure}

RP is one of the most common image padding methods, which can restore input image to the original image. Specifically, it pads the input image using the reflection of the input data, i.e., if we want to add a new row at the top of the input image by RP, the row is obtained from the second row in the input image through vertical reflection and diagonal reflection. Fig.~\ref{fig_reflect_padding} is an example to explain how RP works. For example, the first and the last element in first green row is 0.7 by diagonal reflection. The middle elements are equal to those in second row in blue part by vertical reflection.
\subsection{Batch Normalization (BN)}
In HSI denoising, a CNN can work well if the BN layer can be jointly used~\cite{chang2018hsi}. Particularly, the BN layer allows a higher learning rate, significantly increases the speed of training, and avoids the gradient vanishing or divergence issue. The BN layer begins to work by performing BN when a batch of data is input to the CNN, where BN is achieved by calculating the mean and the standard deviation (std) of the input data~\cite{ioffe2015batch}.

\section{The proposed algorithm}
\label{proposed_algorithm}
\subsection{Algorithm Overview}
\label{algorithm_overview}
Algorithm~\ref{alg_the_proposed_algorithm} shows the framework of the proposed Evolve-CNN method, where the contributions are highlighted in bold and italic. Firstly, the population is randomly initialized, and each individual in the population is generated randomly with the proposed gene encoding strategy (line~\ref{alg_framework_line_1}). Then, the initialized population is evaluated for the parent selection (line~\ref{alg_framework_line_2}). After that, the evolution begins to take effect until the stopping criterion is satisfied (lines~\ref{alg_framework_start_while}-\ref{alg_framework_end_while}). Finally, the expected CNN, built by decoding the selected best individual, is ready for the final training (line~\ref{alg_framework_line_8}).

During the evolution, all of the initialized individuals’ fitness are evaluated. To speed up the evolution, we let individuals be trained only one epoch on the training dataset, while the fitness evaluation is performed on the evaluation dataset. And then, two parent individuals are chosen by the improved slack binary operation from the population, and the offspring is generated with the proposed genetic operation (line~\ref{alg_framework_line_5}). Subsection~\ref{offspring_generation} will illustrate this operation in detail. After offspring generation and fitness evaluation (line~\ref{alg_framework_line_6}), the environment selection using the elite selection mechanism starts to choose the next generation from the existing individuals and the newly generated offspring (line~\ref{alg_framework_line_7}). Then the next generation continue the next round of evolution.

\begin{algorithm}
	\caption{Framework of Evolve-CNN}
	\label{alg_the_proposed_algorithm}
	$P_0 \leftarrow$ Randomly initialize the population with \emph{\textbf{\textit{the proposed gene encoding strategy}}};\\
	\label{alg_framework_line_1}
	Evaluate fitness of $P_0$;\\
	\label{alg_framework_line_2}
	$t \leftarrow$ $0$;\\
	\label{alg_framework_line_3}
	\While{stopping criterion is not satisfied}
	{\label{alg_framework_start_while}
		$t\leftarrow t+1$;\\
		\label{alg_framework_line_4}
		$O_{t}\leftarrow$ Choose parent individual by \textbf{\textit{\textit{the improved slack binary operation}}} from $P_{t-1}$ to generate the offspring with \emph{\textbf{\textit{the proposed genetic operation}}};\\
		\label{alg_framework_line_5}
		Evaluate fitness of $O_{t}$;\\
		\label{alg_framework_line_6}
		$P_{t}\leftarrow$ Environment selection from $P_{t-1}\cup O_{t}$;\\
		\label{alg_framework_line_7}
	}
	\label{alg_framework_end_while}
	\textbf{Return} $P_{t}$.
	\label{alg_framework_line_8}
\end{algorithm}

\subsection{Gene Encoding Strategy}
Generally, the optimal architecture of the CNN is difficult to determine without a prior knowledge. The architecture of the CNN, especially the depth of the CNN, plays a decisive role in the performance of CNNs~\cite{simonyan2014very,bengio2011expressive,bengio2013representation,delalleau2011shallow}. In the traditional design of the CNN architectures, the depth is set based on domain expertise, which is not necessarily accurately assigned for every task at hand. In the proposed algorithm, this issue is addressed by the proposed variable-length gene encoding strategy that is able to automatically find the optimal depth without any expertise. In the proposed encoding strategy, the CNN is built by multiple sequential blocks, and each block consists of a convolution layer, a BN layer and a Rectified Linear Unit (ReLU). In addition, we also add a RP layer after the convolution layer whose kernel size unequal to one to avoid the image size diminishing. Considering that the output image is close to the real one, we set the last block as only a convolution layer whose kernel size and out feature map size are unchangeable.

\begin{figure}[htp]
	\centering
	\includegraphics[width=1\columnwidth]{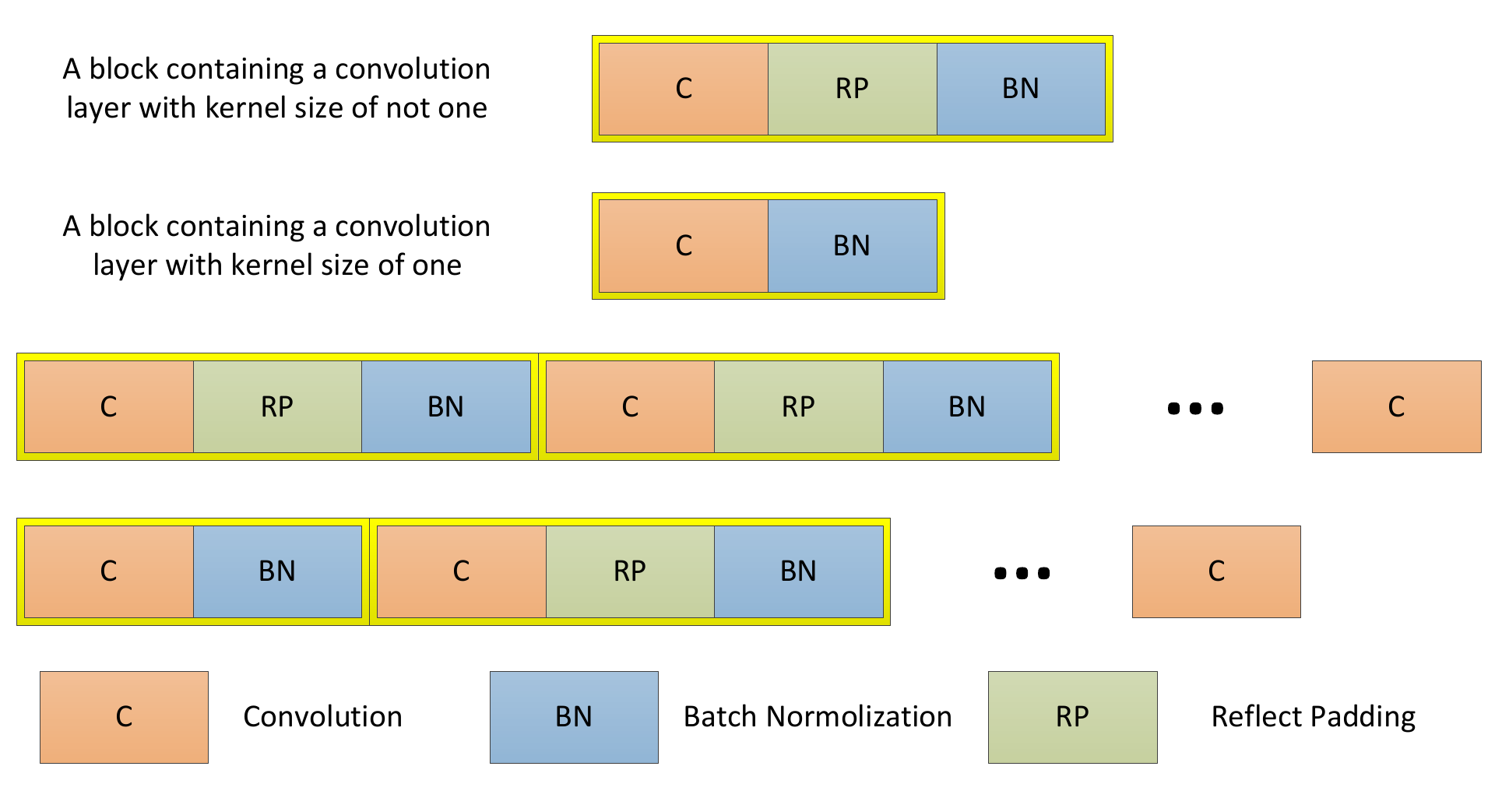}\\
	\caption{An example to show two types of blocks and two chromosomes with different lengths.}\label{fig_dif_block_and_chromosome}
\end{figure}

As mentioned above, three different types of the layers, i.e., the convolution layer, the BN layer, and the RP layer exist in the architectures of Evolve-CNN. Because the RP layer can be regarded as the appurtenance of the convolution layer and the BN layer works well in the default setting, we use the information of the convolution layer to encode the architecture into one chromosome for the evolution. An example of two types of blocks and two chromosomes with different lengths from Evolve-CNN is illustrated in Fig.~\ref{fig_dif_block_and_chromosome}. Commonly, hundreds of thousands of weights may exist in a CNN, which cannot be all initialized explicitly. In the proposed gene encoding strategy, the mean and std are used to efficiently initialize the weights of CNN using the Gaussian initialization mechanism. In addition, the important parameters of the convolution layer, including the filter width, filter height, and the number of feature maps are all encoded into the chromosome.

\begin{algorithm}
	\caption{Population Initialization}
	\label{alg_pop_init}
	\KwIn{the population size $N$; the maximal number of blocks, $N_{max}$; the minimal number of blocks, $N_{min}$}
	\KwOut{Initialized population, $P_0$}
	$P_0\leftarrow\emptyset$ \\
	\label{alg_pop_init_line1}
	\While{$\left| P_0 \right| \leq N$}{
		\label{alg_pop_init_line2}
		$head\leftarrow \emptyset$;\\
		\label{alg_pop_init_line3}
		$r\leftarrow$ uniformly generate a random integer between $[N_{min}, N_{max}]$;\\
		\label{alg_pop_init_line4}
		\While{$\left| head \right| < r$}{
			\label{alg_pop_init_line5}
			$l\leftarrow$ generate a convolution layer with random settings;\\
			$head\leftarrow head\cup l$;\\
			$s\leftarrow$the kernel size of the convolution layer;\\
			\If{$s\neq 1$}{
				$l\leftarrow$ generate a reflect padding layer with $\frac{s-1}{2}$ size;\\
				$head\leftarrow head\cup l$;\\
			}
			$l\leftarrow$ generate a batch-normalization layer with random settings;\\
			$head\leftarrow head\cup l$;\\
		}
		\label{alg_pop_init_line15}
		$l\leftarrow$ generate a convolution layer with predefined number of feature maps and kernel size;\\
		\label{alg_pop_init_line16}
		$head\leftarrow head\cup l$;\\
		\label{alg_pop_init_line18}
		$P_0\leftarrow P_0\cup head$;
	}\label{alg_pop_init_line19}
	\Return $P_0$
	\label{alg_pop_init_line20}
\end{algorithm}

Algorithm~\ref{alg_pop_init} shows the population initialization by using the proposed genetic encoding strategy, where $\left| \cdot \right|$ is a cardinality operator. Firstly, an empty population of size N is initialized (line~\ref{alg_pop_init_line1}). Secondly, individuals are created randomly to fill in the population (lines~\ref{alg_pop_init_line2}-\ref{alg_pop_init_line19}) until the population reached its predefined size of N. Finally, an initialized population $P_0$ is returned (line~\ref{alg_pop_init_line20}). The CNN architecture encoded by the individuals is divided into two parts, namely the head and the last convolution layer, according to whether there are fixed parameters or not, and the individual initialization is also divided into two parts. The first part is generating the head with the random settings (lines~\ref{alg_pop_init_line3}-\ref{alg_pop_init_line15}), and the second part is adding a partially randomly initialized convolution layer (lines~\ref{alg_pop_init_line16}-\ref{alg_pop_init_line18}).

Noting that line~\ref{alg_pop_init_line4} sets the upper and the lower bounds of the depth because a shallow CNN may not perform well on the HSI denoising and a deep CNN will take up a lot of computing resources unnecessarily. In addition, because the output size of the CNN is determined, and the feature map size and the filter size of the last convolution layer are fixed accordingly, while the mean and std of filter elements are variable (line~\ref{alg_pop_init_line16}). In order to find a suitable RP layer, the kernel size of the convolution layer must be an odd number.

\begin{figure*}[htp]
	\centering
	\includegraphics[width=2\columnwidth]{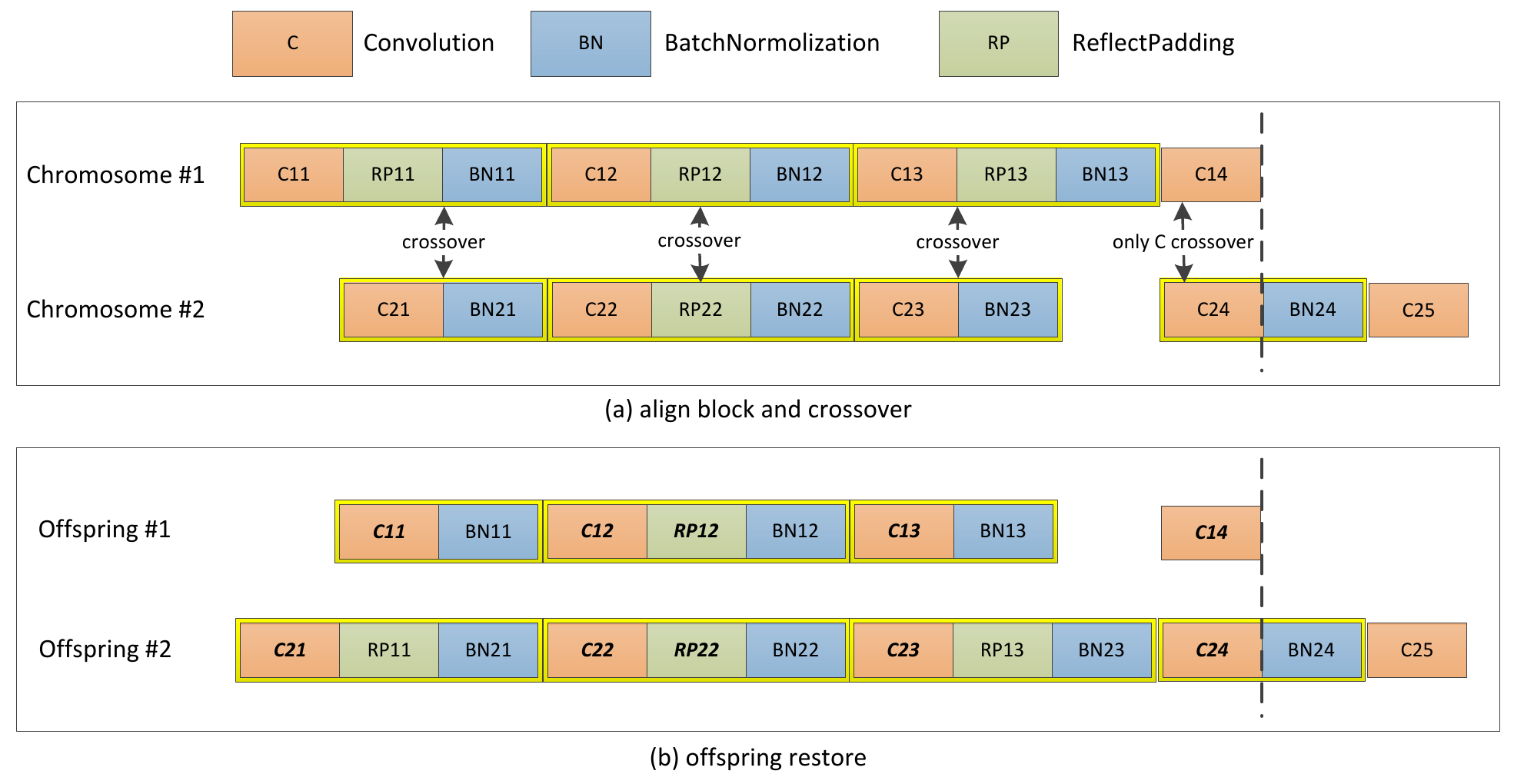}\\
	\caption{A crossover example.}\label{fig_recombinate_operation}
\end{figure*}

\subsection{Offspring Generation}
\label{offspring_generation}
Because the Mean Squared Error (MSE) represents the pixel value difference between the denoised image and the original image, we use it as the fitness evaluation criterion for the task investigated in this work. Specifically, the MSE is calculated by Equation~(\ref{equ_mse_calculation})
\begin{equation}
\label{equ_mse_calculation}
\text {MSE}=\dfrac{||D-O||^2}{M\times N}
\end{equation}
where $||\cdot||$ denotes the L2 norm of a matrix, D and O denote the pixel values of the image after the denoising and the original image, respectively, M and N denote the length and width of the image. The smaller MSE is, the better denoising effect is, and the better fitness degree the individual is. In practice, there will be multiple CNN architectures resulting in almost the same MSE, while the one having the least number of parameters should be the promising one because fewer parameters of the CNN means the low complexity and potentially provides better generalization ability. Thus, the number of CNN parameters, indicating the complexity of the CNN, is also used as part of the individuals’ fitness.

Just like traditional GAs and as mentioned in Section~\ref{algorithm_overview}, after the population initialization and the fitness evaluation for the original population, the offspring will be generated with the proposed genetic operations. Traditionally, the offspring are generated by performing the genetic operators, which usually consists of the crossover operator and the mutation operator. The steps of generating the offspring are shown below:

\begin{enumerate}
\item Select two parents by Algorithm~\ref{alg_sbts} (Section~\ref{environmentail_selection} will illustrate this algorithm in detailed); 
\item Perform mutation and crossover on the chosen parent individuals, and generate the offspring;
\item Store the offspring and reperform Step 1 until the number of offspring reaches the predefined size.
\end{enumerate}

In the proposed algorithm, the mutation operation is divided into two steps. The first step allows the encoded variable information mentioned above to mutate in a given range by the Polynomial Mutation operator (PM)~\cite{deb2001multi}. The second step is to randomly increase or decrease the depth of the CNN. This step may perform on each position of the CNN, where the position is randomly decided. After that, one particular mutation operation is randomly selected from: 1) adding a block which is randomly initialized, 2) removing a block with the exception of the last layer, and 3) keeping the depth unchanged. Then, the operation performs on the selected position. Each type of the operation has a 1/3 chance of being randomly selected.

In the proposed crossover operation, we use the Simulated Binary Crossover (SBX)~\cite{deb1995simulated} to perform the crossover owing to its promising performance on the real numbers, i.e., the encoded variable information in the proposed encoding strategy.

Fig.~\ref{fig_recombinate_operation} illustrates the crossover process where C denotes the convolution layer, BN denotes the BN layer, and RP denotes the RP layer. In this example, the first chromosome has three blocks while the second chromosome has four. Initially, the blocks of each chromosome align based on the order as shown in Fig.~\ref{fig_recombinate_operation}.a. Secondly, the matched blocks perform the crossover operation. There are two different situations when doing the crossover operation: one is for the same structures when both blocks consist of C+RP+BN or C+BN. The other is for the different structures when one block is C+RP+BN and the other is C+BN. The first case perform the crossover operation by just swapping the encoded variable information, while for the second case, not only the encoded variable information, but also taking the RP to the side having no RP. It’s worth noting that if the lengths of the chromosomes are not the same, the crossover operation is performed on the last layer of the shorter chromosome and the corresponding convolution layer of the longer one. In order to get the valid image size, the number of the feature maps will not participate this operation at the last layer. Finally, the offspring is generated as shown in Fig.~\ref{fig_recombinate_operation}.b. where bold and italics indicate that this layer has been modified.

\begin{algorithm}
	\caption{Slack Binary Tournament Selection}
	\label{alg_sbts}
	\KwIn{MSE threshold, $\alpha$; complexity threshold, $\beta$; The population}
	\KwOut{The selected individual}
	Randomly select two individuals from the population\\
	\label{alg_sbts_line1}
	$i_1\leftarrow$ the individual with smaller MSE;\\
	\label{alg_sbts_line2}
	$i_2\leftarrow$ the other individual;\\
	\label{alg_sbts_line3}
	$m_1,m_2\leftarrow$ the MSE of $i_1,i_2$;\\
	\label{alg_sbts_line4}
	$c_1,c_2\leftarrow$ the complexity of $i_1,i_2$;\\
	\label{alg_sbts_line5}
	\uIf{$\dfrac{m_2-m_1}{m_1} > \alpha$}{
		\label{alg_sbts_line6}
		\Return $i_1$\\
		\label{alg_sbts_line7}
	}
	\Else{
		\uIf{$c_1-c_2 > \beta$}{
			\label{alg_sbts_line9}
			\Return $i_2$\\
			\label{alg_sbts_line10}
		}
		\Else{
			\label{alg_sbts_line11}
			\Return $i_1$\\
			\label{alg_sbts_line12}
		}
	}
\end{algorithm}

\subsection{Environmental Selection}
\label{environmentail_selection}
In order to maintain a population with the promising convergence and diversity, the elite mechanism is used in the developed environment selection which is named as “Slack Binary Tournament Selection”. Algorithm~\ref{alg_sbts} shows the details of the design. Firstly, two individuals are selected randomly from the population (line~\ref{alg_sbts_line1}). Secondly, the individual with the smaller MSE assigns to $i_1$, and the other one assigns to $i_2$ (lines~\ref{alg_sbts_line2}-\ref{alg_sbts_line3}). Thirdly, their MSE values and complexity assign to the corresponding parameters (lines~\ref{alg_sbts_line4}-\ref{alg_sbts_line5}). In the end, the proportion (line~\ref{alg_sbts_line6}) and the difference (line~\ref{alg_sbts_line9}) are compared with the given threshold to choose the better individual. Specifically, if a CNN’s performance is much better than the other one’s (lines~\ref{alg_sbts_line6}-\ref{alg_sbts_line7}), we choose the better one directly, when their performances are about the same, picking the one with less complexity (lines~\ref{alg_sbts_line9}-\ref{alg_sbts_line10}). When the performance and the complexity are both about the same, the performance is preferred (lines~\ref{alg_sbts_line11}-\ref{alg_sbts_line12}).

Noting that we use a proportion instead of a simple difference as a threshold, because in the earlier stage of the evolution, the MSE value of all individuals is relatively large, but in the later stage of the evolution, most individuals have good denoising performance, i.e., the MSE value is generally small. So the proportion is better to maintain the consistency of selection operation. When the evolution process is finished, the best individual which has the smallest MSE is selected from the last generation to perform the final training of the CNN.

\section{EXPERIMENT DESIGN}
\label{experiments}
In order to quantify the performance of the proposed Evolve-CNN algorithm, an experiment is designed to compare with the state-of-the-art peer competitors on the chosen benchmark dataset. In the following, the benchmark dataset and how to build the training, the evaluation, and the test dataset are introduced at first. Then, the peer competitors are listed. After that, the parameter settings, including the training parameters for the proposed algorithm, are detailed.

\subsection{Benchmark Dataset}
The Indian Pines dataset~\cite{aviris2012indiana} which is widely used in HSI denoising~\cite{chang2018hsi,zhang2019hybrid} is chosen as the benchmark dataset. The dataset is gathered by AVIRIS sensor over the Indian Pines test site in North-western Indiana. Particularly, the Indian Pines dataset contains two HSIs with different sizes. The first is with the size of 614$\times$2678$\times$220, and the second is with the size of 1848$\times$614$\times$220. In order to do a fair comparison, the Gaussian noise whose intensity level $\sigma=0.778$ is added to the original images to generate the data used by the compared algorithms for the denoising.

Based on the conventions, the dataset has been divided into three parts for the experiments, i.e., the training dataset, the evaluation dataset, and the test dataset, which are randomly selected and have 17535, 4000 and 4838 images, respectively, i.e., account for 66.5\%, 15.2\%, and 18.3\%, respectively. In order to obtain the sufficient images to train the CNN, the original clean images and generated noise images are cropped with the size of 30$\times$30, with the sampling stride equaling to 10.

\subsection{Peer Competitors}
Multiple state-of-the-art HSI denoising methods are chosen as the peer competitors. Considering the proposed algorithm focusing on the CNN methods, we also choose the algorithm proposed by Chang~\cite{chang2018hsi}, which recently reported its promising performance on HSI denoising, as one of the peer competitors in this experiment. 

Overall, the chosen peer competitors are listed below: 1) the BM3D method~\cite{dabov2007image}; 2) the low-rank tensor approximation (LRTA) method~\cite{renard2008denoising}; 3) the Total-variation-regularized low-rank matrix factorization (LRTV) method~\cite{he2015total}; 4) the tensor dictionary learning method (TDL)~\cite{peng2014decomposable}; 5) the block matching 4-D filtering method (BM4D)~\cite{maggioni2012nonlocal}; 6) the intrinsic tensor sparsity regularization method (ITSReg)~\cite{xie2016multispectral}; 7) the LRMR method~\cite{zhang2013hyperspectral}; 8) the Chang’s method (Artificial-CNN)~\cite{chang2018hsi}.

Noting that, the optimal CNN selected by the proposal algorithm is chosen based on the evaluation dataset after it has been trained on the training dataset. When the evolutionary process is finished, the best performance is obtained by training it on both the training dataset and the evaluation dataset following the conventions of the deep learning community. The test dataset remains unseen during this process.

\subsection{Parameter Settings}
In the experiment, the parameter settings include the parameters of the evolution in the proposed algorithm and the parameters of the peer competitors.

All the parameter settings of the evolution are specified following the conventions of GA community~\cite{deb1995simulated,deb2001multi}. Considering both the performance and the complexity, the depth of CNN varies from 4 to 8. The filter size of convolution layer, equivalent to the filter width and filter height, are randomly chosen from $\left\{1, 3\right\}$. All layer’s feature map sizes vary from 128 to 512 except the last convolution layer whose size is fixed. Preliminary experiment and experience specify the mean range from -0.8 to 0.8, and the std from 0 to 0.5. In addition, the population size and the total generation number are set to be 30 and 10, respectively. The distribution index of the SBX and PM are both set to 1 based on the conventions~\cite{deb1995simulated,deb2001multi}, and their associated probabilities are specified as 0.9 and
0.2, respectively. In the environment selection, the elitism rate is specified as 20\% based on the Pareto principle. As mentioned in Section~\ref{offspring_generation}, the MSE and the complexity are used in the fitness function during the evolutionary stage.

In addition, all the parameters of peer methods are set by default because their default settings can give them the best performance.

\subsection{Training Details}
In this subsection, we give the training details in both the evolution process and the final train.

In the evolution stage, each individual needs to be trained at first for the fitness evaluation. We use the Adam method~\cite{kingma2014adam} as the optimizer for this stage, and the learning rate is set to 0.004 and exponential decay rate is set as the default setting. The training dataset will not be shuffled at evolution stage to ensure the consistency in training each individual.

For the final train, the Adam configured with default setting is used again as the optimizer of the CNN, respectively and the learning rate is initialized to 0.001. It is worth noting that, the training dataset and the evaluation dataset are both used to train the CNN, and these two datasets are shuffled in this stage.

The training of the Artificial-CNN uses the same settings as those of the proposed Evolve-CNN, in addition to its weight initialization method that uses the Xavier initializer~\cite{glorot2010understanding}. We employed the Pytorch to implement both CNNs on a Ubuntu server with an NVIDIA 2080 Ti GPU card. 

Noting that, the batch sizes of the proposed evolve-CNN method and the Artificial-CNN are set to be 100 for both the training and the test stage.  Furthermore, we did not limit the training epoch but stop the training and test by investigating the performance does not increase.

\begin{table*}[htp]
	\renewcommand{\arraystretch}{1.3}
	\caption{Average Denoising Performance Comparison Eight Competing Methods With Respect To Four PQIs of The INDIAN PINE Under Gaussian Noise $\sigma=0.778$}
	\label{tab_overall_result}
	\begin{center}	
		\begin{tabular}{|l|c|c|c|c|c|c|c|c|c|c|}
			\hline 
			\diagbox{Measure}{Method} & Nosiy & BM3D & LRTA & LRTV & TDL & BM4D & ITSReg & LRMR & Artificial-CNN & Evolve-CNN \\
			\hline 
			MPSNR & 50.311 & 62.881 & 64.939 & 64.852 & 67.248 & 65.122 & 66.979 & 65.115 & 69.317 & \textbf{70.051}\\ 
			\hline 
			MSSIM & 0.98997 & 0.99905 & 0.99937 & 0.99954 & 0.99967 & 0.99951 & 0.99956 & 0.99958 & 0.99974 & \textbf{0.99977}\\ 
			\hline 
			MFSIM & 0.96443 & 0.97201 & 0.99030 & 0.98072 & 0.98883 & 0.98107 & 0.98829 & 0.98599 & 0.99197 & \textbf{0.99213}\\ 
			\hline 
			MERGA & 29.857 & 7.567 & 5.877 & 5.929 & 4.483 & 5.575 & 4.906 & 5.333 & 4.095 & \textbf{3.976}\\ 
			\hline 
		\end{tabular} 
	\end{center}
\end{table*}

\begin{figure*}[htp]
	\centering
	\includegraphics[width=1.8\columnwidth]{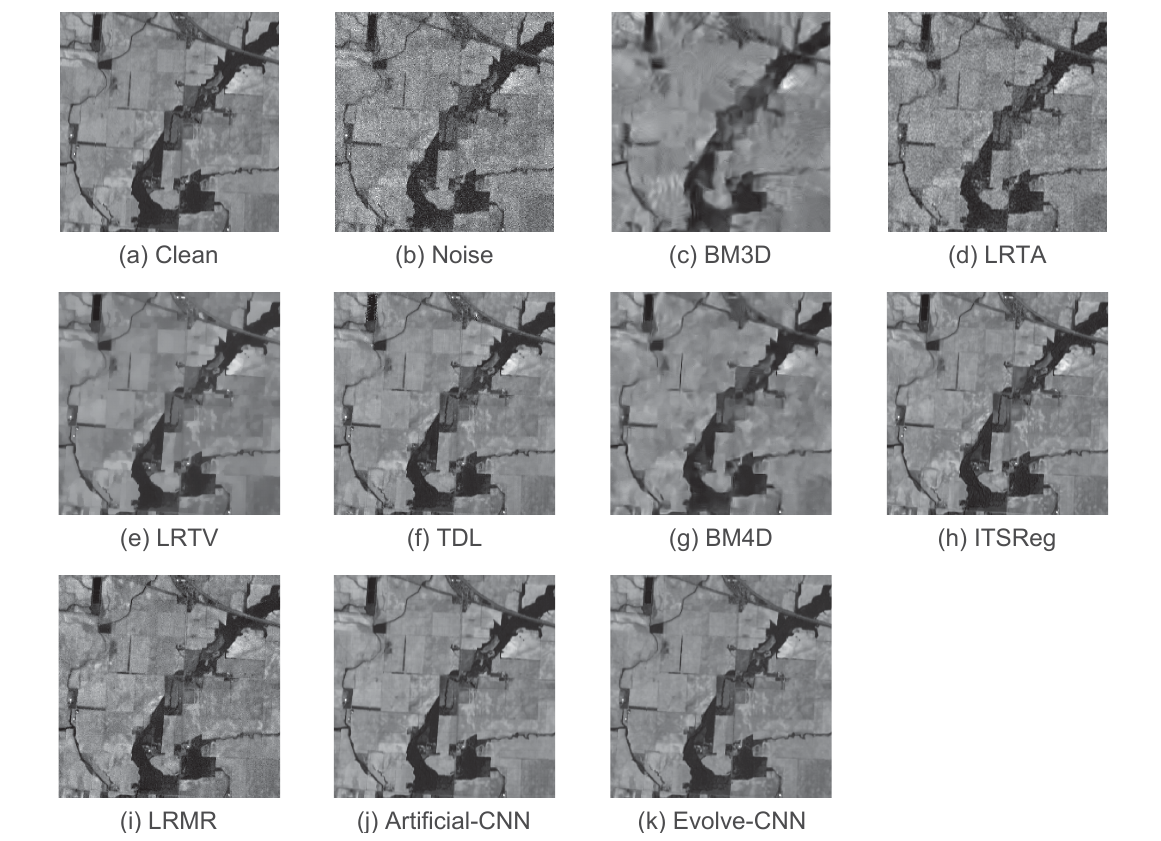}\\
	\caption{Visual comparison in band 187. (a)Clean. (b)Noise. (c)BM3D. (d)LRTA. (e)LRTV. (f)TDL. (g)BM4D. (h)ITSReg. (i)LRMR. (j)Artificial-CNN. (k)Evolve-CNN.}. \label{fig_visual_comparison}
\end{figure*}

\section{Experimental Results and Analysis}
\label{result}
To acquire an integrated comparison for all peer competitors and the proposed Evolve-CNN method, the quantitative evaluation indicators and a visual comparison are used to analyze the experimental results. We employ four Image Quality Measurements (IQMs) as the quantitative evaluation indicators, including the Mean Peak Signal to Noise Ratio (MPSNR), the Mean Structural SIMilarity index (MSSIM)~\cite{wang2004image}, the Mean Feature Similarity index (MFSIM)~\cite{zhang2011fsim}, and the Mean Erreur Relative Globale Adimensionnelle de Synthèse (MERGA)~\cite{wald2002data}. MPSNR, MSSIM and MFSIM evaluate the similarity between the target image and the reference image based on the MSE value, the structural consistency and the perceptual consistency. The larger they are, the more similar the two images are, i.e., the better the method is. Different from the former three indicators, MERGA measures the fidelity, and the smaller MERGA is, the better the method is.

\subsection{Overall Results}
\label{overall_results}
The average results of the four performance evaluation indicators are listed in Table~\ref{tab_overall_result}. Band 187 was chosen randomly for visual comparison that is shown in Fig.~\ref{fig_visual_comparison}. 

In Table~\ref{tab_overall_result}, the best performance of each evaluation indicator is marked in bold. As can be seen from Table~\ref{tab_overall_result}, the CNN-based methods have the absolute leading position among the compared algorithms. Specifically, the proposed Evolve-CNN  provides the highest values in MPSNR, MSSIM, and MFSIM, and the lowest MERGA, and becomes the only method with MPSNR over 70. In Fig.~\ref{fig_visual_comparison}, the BM3D could not deal with this noise well, furthermore, the BM4D produces over-smoothing in this result. Residual gaussian noise is clearly visible in LRTA and LRMR, and the left methods have high fidelity.

\subsection{Comparisons with Artificial-CNN}
The CNN-based methods are compared separately to tell the effect from the automatic architecture selection mechanism. Table~\ref{tab_Comparision_with_ACNN} shows the depth, the number of total parameters, the training time and the performance derived from evaluation indicators. We define one hour on one GPU service as a GPUh to measure the training time.

As can be seen from Table~\ref{tab_Comparision_with_ACNN}, the Artificial-CNN method is much more complicated than the proposed Evolve-CNN method. Meanwhile, the training time of Artificial-CNN is three times more than that of Evolve-CNN. However, the shorter CNN perform better in the evaluation indicators, although both CNNs have excellent denoising effect. Furthermore, in terms of the IQMs results upon the four employed measures, the Evolve-CNN model is twice significantly better than Artificial-CNN, and twice significantly equal to Artificial-CNN. 

\begin{table}[htp]
	\renewcommand{\arraystretch}{1.3}
	\caption{Comparison With Artificial CNN. The symbols ``+'', ``=,'' and ``-'' denote whether the IQMs results of the proposed Evolve-CNN are better than, equal to or worse than that of the Artificial-CNN with a significant level $1\%$}
	\label{tab_Comparision_with_ACNN}
	\begin{center}
		\begin{tabular}{p{0.3\columnwidth}<{\centering}|p{0.25\columnwidth}<{\centering}|p{0.25\columnwidth}<{\centering}}
			\hline
			& Evolve-CNN & Artificial-CNN \\
			\hline
			depth & 5 &  16\\
			total params & 1,838,603 &  32,254,420\\
			training time & 15 GPUhs & 54 GPUhs\\
			evolutionary time & 0.5 GPUh & 0\\
			+/=/- &  & 2/2/0\\
			\hline
		\end{tabular}
	\end{center}
\end{table}

\section{Conclusions}
\label{conclusion}
We have proposed an automatic method by using the genetic algorithm to design the CNNs for HIS denoising. Particularly, we have designed an improved genetic encoding strategy for encoding the CNN architectures and the weight initialization parameters, the corresponding genetic operators to effectively and efficiently find the optimal CNN architecture during the evolution process, and a new tournament selection to choose the promising parent individuals for the evolution performance. In the experiment, we found the proposed method is much better than other methods in terms of the evaluation indicator values and the visual assessments. The automated CNN by the proposed algorithm has much fewer parameters than the state-of-the-art CNN peer competitors. Furthermore, the automated CNN by our proposed algorithm can achieve the performance of that designed by experts. In the future, we will devote to the research of fitness evaluation methods, and also  investigate which new components contribute more to the performance.



\bibliographystyle{IEEEtran}
\bibliography{IEEEabrv,mybibfile}

\end{document}